\documentclass[10pt,twocolumn,letterpaper]{article}

\usepackage{cvpr}              

\usepackage{booktabs}
\usepackage{multirow}
\usepackage{graphicx}
\usepackage{subcaption}
\usepackage{float}
\usepackage[normalem]{ulem}
\usepackage[table]{xcolor}
\useunder{\uline}{\ul}{}

\definecolor{cvprblue}{rgb}{0.21,0.49,0.74}
\usepackage[pagebackref,breaklinks,colorlinks,allcolors=cvprblue]{hyperref}

\definecolor{myRed}{rgb}{0.8, 0.2, 0.2}
\definecolor{myOrange}{rgb}{0.8, 0.45, 0.0}
\definecolor{myBlue}{rgb}{.0, .0, 1.0}

\def\confName{CVPR}

\usepackage{multirow}
\usepackage{bm}
\title{DeX-Portrait: Disentangled and Expressive Portrait Animation\\via Explicit and Latent Motion Representations
}

\author{
Yuxiang Shi\textsuperscript{1,*,\textdagger}
\space\space
Zhe Li\textsuperscript{2,*}
\space\space
Yanwen Wang\textsuperscript{3,\textdagger}
\space\space
Hao Zhu\textsuperscript{3}
\space\space
Xun Cao\textsuperscript{3}
\space\space
Ligang Liu\textsuperscript{1}
\\
\small{\textnormal{
\textsuperscript{1}University of Science and Technology of China
\space\space\space\space\space\space 
\textsuperscript{2}Central Media Technology Institute, Huawei
\space\space\space\space\space\space 
\textsuperscript{3}Nanjing University
}}
}

\begin{document}

\twocolumn[{
    \maketitle
    \vspace{-0.35in}
    \begin{center}
        \centering
        \includegraphics[width=1\textwidth]{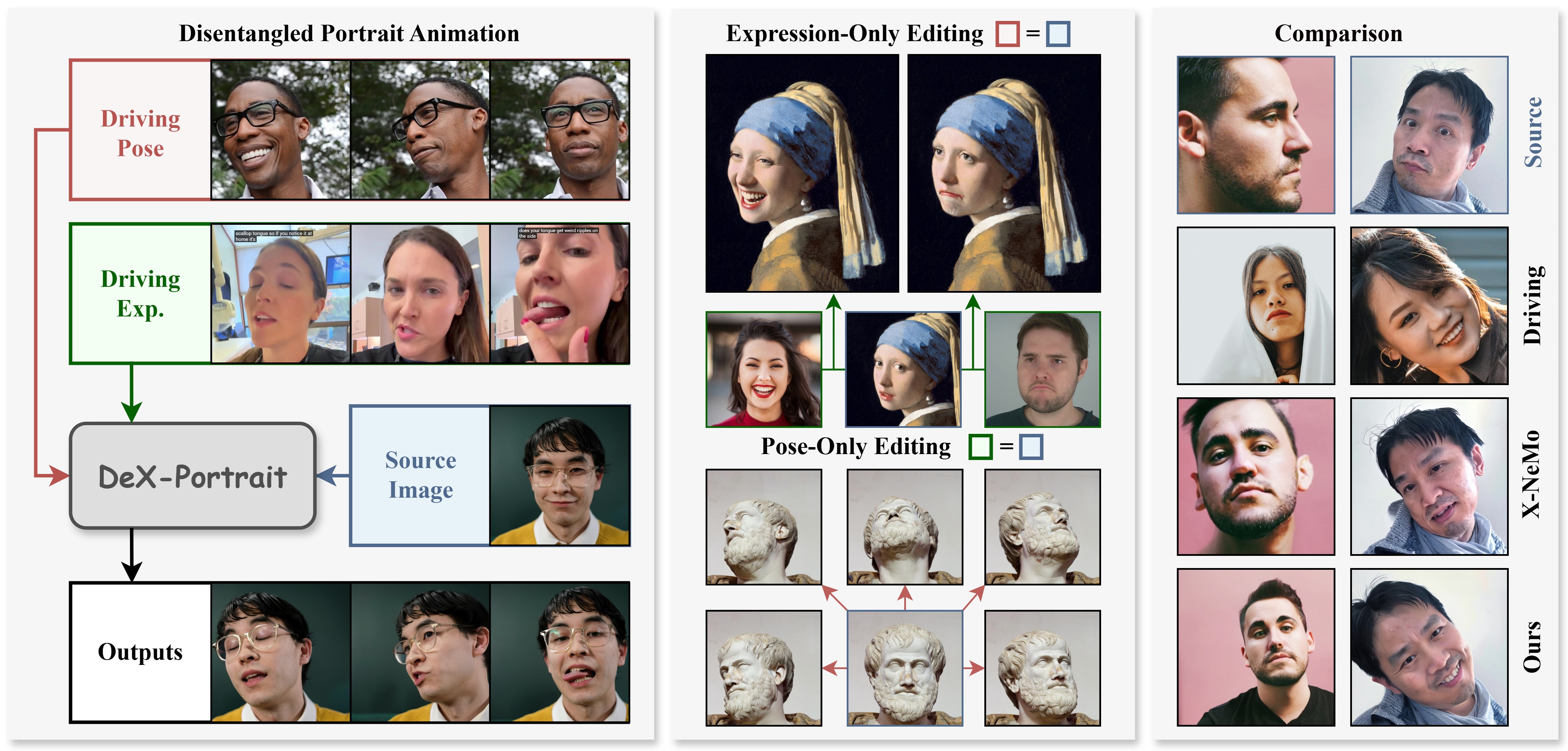}
        \captionof{figure}{
        \textbf{Left:} Given an arbitrary source portrait image, driving expression images, and driving pose images, DeX-Portrait achieves disentangled and expressive portrait animation. 
        \textbf{Middle:} DeX-Portrait enables expression-only and pose-only editing while keeping the other driving signal exactly the same as the source. 
        \textbf{Right:} Compared with the state-of-the-art work, X-NeMo \cite{xnemo}, our method offers superior fine-grained control over head pose including rotation, translation and scale.
        }
        \label{fig:teaser}
    \end{center}
}]
\let\thefootnote\relax\footnotetext{*Indicates equal contribution to this work.}
\let\thefootnote\relax\footnotetext{\textdagger Work done during an internship at Huawei.}
\begin{abstract}
Portrait animation from a single source image and a driving video is a long-standing problem.
Recent approaches tend to adopt diffusion-based image/video generation models for realistic and expressive animation.
However, none of these diffusion models realizes high-fidelity disentangled control between the head pose and facial expression, hindering applications like expression-only or pose-only editing and animation.
To address this, we propose DeX-Portrait, a novel approach capable of generating expressive portrait animation driven by disentangled pose and expression signals.
Specifically, we represent the pose as an explicit global transformation and the expression as an implicit latent code.
First, we design a powerful motion trainer to learn both pose and expression encoders for extracting precise and decomposed driving signals.
Then we propose to inject the pose transformation into the diffusion model through a dual-branch conditioning mechanism, and the expression latent through cross attention.
Finally, we design a progressive hybrid classifier-free guidance for more faithful identity consistency.
Experiments show that our method outperforms state-of-the-art baselines on both animation quality and disentangled controllability. Project page is avaliable at \href{https://syx132.github.io/DeX-Portrait/}{https://syx132.github.io/DeX-Portrait/}
\end{abstract}   
\section{Introduction}
\label{sec:intro}
One-shot portrait animation, aiming at animating a source portrait image using a driving video, has been a popular topic due to its value in digital content creation.
With the recent development of diffusion-based image/video generation models \cite{wan, flux, SD1_5}, researchers tend to modify and finetune them for expressive portrait animation \cite{xnemo, wananimate}.
Although these SOTA diffusion-based approaches realize high-quality facial animation, they still suffer from the controllability, especially individual controls on the head pose and facial expression, hindering the applications like expression-only or pose-only editing.



To achieve disentangled pose and expression controls, a plausible way \cite{mu2025flap, wang2025dreamactor-h1} is to represent the portrait motion with the head pose and expression blendshapes of 3D Morphable Models (3DMM) \cite{li2017flame, wang2022faceverse, gao2024portrait}.
However, the performance of these methods is limited by the accuracy of 3DMM trackers and the representation ability of blendshapes.
Consequently, they struggle to capture complex and subtle facial motions like sticking out the tongue and frowning. 
On the other hand, the state-of-the-art approach, X-NeMo \cite{xnemo}, encodes the portrait motion as a 1D latent code capable of capturing expressive facial expression.
Unfortunately, the latent code entangles the pose and expression and fails to precisely control the head rotation, scale and translation as shown in \cref{fig:teaser}.

To this end, we propose \textit{DeX-Portrait}, a diffusion-based framework that leverages explicit and latent motion representations for both disentangled and expressive portrait animation.
Specifically, the head pose is represented as an explicit global transformation including a rotation, translation and scale (RTS), while the facial expression is represented as a latent code.
The first challenge lies in disentangling the pose and expression encoders for extracting expression-agnostic pose transformation and pose-agnostic expression code.
Thus we design a powerful GAN-based motion trainer (\cref{fig:pipeline} (a)) by firstly applying 3D warping using a RTS-derived transformation and then modulating the generator using the expression code through Adaptive Instance Normalization (AdaIN) \cite{huang2017adain}.
To prevent the pose leakage from the expression encoder, we design a series of augmentation strategies such as central cropping, random rotation and cross-view driving.

With the disentangled pose and expression encoders obtained, the second challenge lies in how to effectively inject the pose and expression signals into the diffusion model.
We propose a novel dual-branch pose conditioning mechanism for precise pose control.
As shown in \cref{fig:pipeline}, In the first branch, we map the pose RTS into a ray map and concatenate it with the noisy latent. 
In the other branch, we warp the intermediate features of the source portrait images through a 3D warping module and concatenate them with the corresponding features in the denoising UNet.
Such a dual-branch pose injection enables the diffusion model to precisely control the head rotation, translation and scale.
The expression code is injected via cross attention \cite{vaswani2017attention} like previous works \cite{xnemo,luo2025dreamactor-m1,fantasyportrait}.
Thanks to the hybrid motion representations and injection methods, our model realizes both expressive and disentangled portrait animation.
In addition, inspired by FLOAT \cite{ki2025float}, we propose a progressive hybrid classifier-free guidance (CFG) in the denoising process by incorporating the pose and expression conditions successively for more stable identity consistency.


In conclusion, our core technical contributions are:
\begin{itemize}
    \item 
    DeX-Portrait, a diffusion-based framework that leverages explicit and latent motion representations for portrait animation, realizing both disentangled and expressive pose and expression controls.
    \item 
    A powerful motion trainer that learns disentangled and precise pose and expression encoders through 3D warping and AdaIN modules.
    \item 
    A dual-branch pose conditioning mechanism that injects the pose transformation into the diffusion model through the 3D warping module and the ray map.
    \item 
    A progressive bybrid CFG that gradually incorporates the expression condition for more consistent identity.
    
\end{itemize}
 
\section{Related Work}
\subsection{Generalizable Portrait Animation}
GAN or diffusion based portrait animation models first encode driving videos into motion representations, which are then used to animate arbitrary source portraits.
Traditional approaches relied on explicit motion representations, such as 3D Morphable Models (3DMM) \cite{zhao2024invertavatar, taubner202cap4d, taubner2025mvp4d, ji2021audio, sun2025vividtalk}, facial landmarks \cite{fye, qiu2025skyreels, wang2019few}, or dense optical flow maps \cite{fomm} to disentangle the appearance and motion. 
While structured representations enable interpretable control, the process of explicit feature extraction inherently introduces biases, leading to limited accuracy in modeling dynamic expressions and poor generalization to large pose variations or complex scenarios.
To address these issues, recent works have shifted to implicit motion representations, embedding motion information directly into latent spaces for end-to-end training \cite{facevid2vid, liveportrait, emoportraits, ji2022eamm}. Among them, \cite{nhr, pdfgc, lia, bai2024universal} regard motion as a style and leverage StyleGAN-like architectures to produce animated results.
Subsequently, diffusion-based models \cite{xnemo, wananimate, luo2025dreamactor-m1, fantasyportrait} have also begun to incorporate such implicit control signals.
However, implicit motion encoders rely on GANs for training, which often leads to entanglement between motion representations and identity features in the latent space.
Different from previous approaches, DeX-Portrait incorporates an implicit expression representation and an explicit head pose representation. Via a novel motion injection method, our model can maximally mitigate the identity leakage issue while ensuring high-fidelity facial expression generation results.

\begin{figure*}[htbp]
    \centering  
    \includegraphics[width=\linewidth]{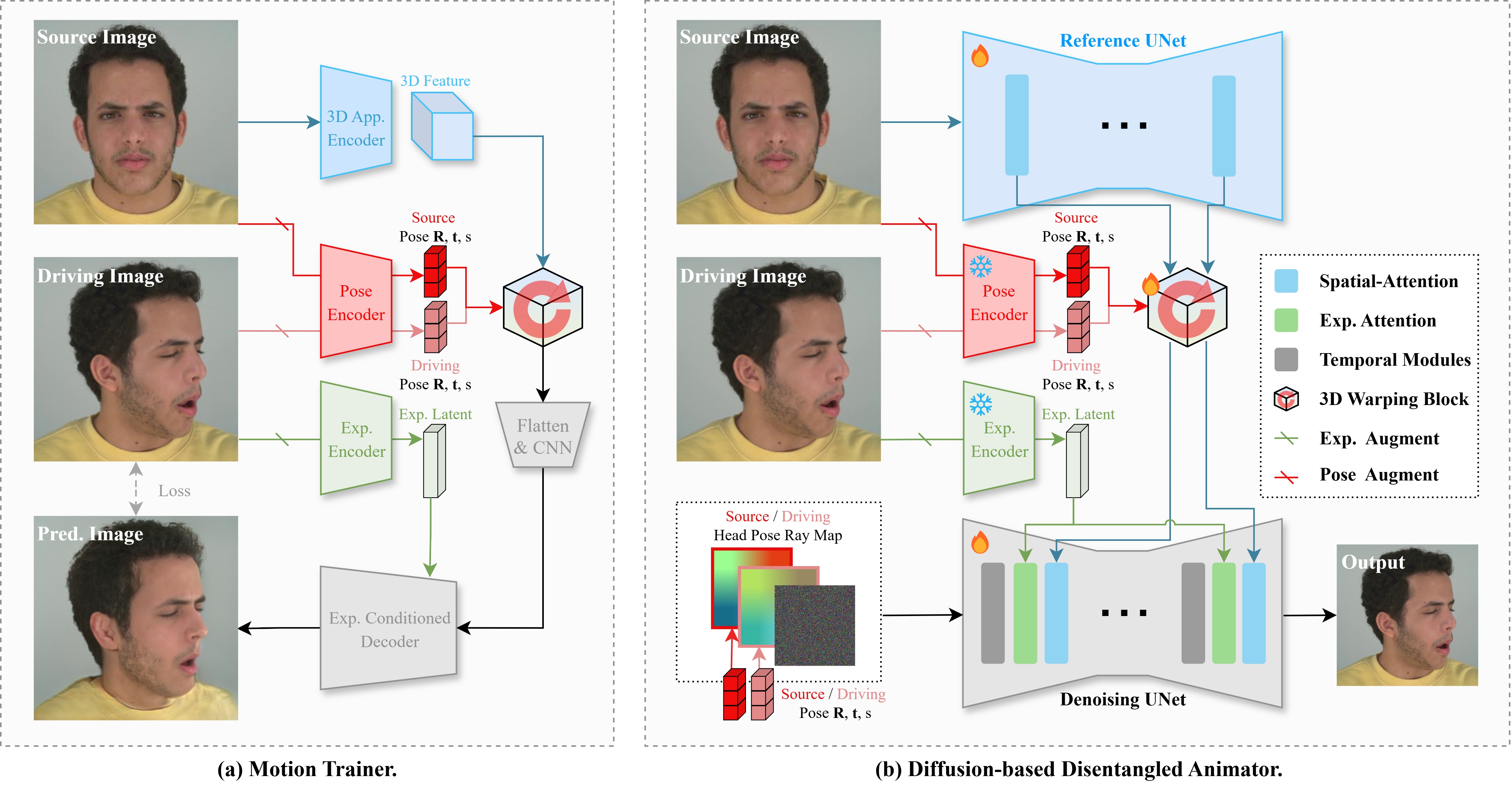}
    \caption{Our pipeline consists of two stages: (a) Training a disentangled pose and expression encoder using a motion trainer. (b) Taming a latent diffusion model for disentangled and expressive portrait animation.}
    \label{fig:pipeline}
\end{figure*}

\subsection{Disentanglement}
Recent years, several portrait animation models aim for disentangled control of facial expressions and head posesfor diverse applications. These methods can be categorized into three classes: The first class leverages manually defined facial models (e.g., 3DMM) to extract expression/pose parameters of the driving face \cite{mu2025flap, wang2021neural, zhang2023sadtalker}, and achieves disentangled control by replacing these parameters during inference. The second class relies on semantic extractors (e.g., action units, CLIP) for explicit disentangled control \cite{wei2025magicface, hellomeme, chang2025talkingheadgenerationauguided}. Both suffer from limited feature extraction accuracy and often fail to faithfully replicate the expressions of the driving subject. The third type employs latent pose and expression features \cite{liveportrait, emoportraits, pang2023dpe, drobyshev2022megaportraits, pdfgc}, guiding feature extractors to extract mutually disentangled expression and pose features via physically interpretable network architectures, loss functions, or other approaches. Such methods have significantly improved generation accuracy; unfortunately, they can only be end-to-end trained based on GANs, limiting output fidelity. Our model incorporates a unique GAN-based motion trainer to train high-precision disentangled latent expression and pose features, and integrates a Diffusion-based generator to achieve high-fidelity results.

\section{Method}
\label{sec:method}

Given a source portrait image $\mathbf{I}_\text{s}$ and a driving sequence $\{\mathbf{I}_\text{d}\}$, our objective is to generate a portrait animation sequence $\{\hat{\mathbf{I}}(\text{ID}_\text{s}, \text{pose}_\text{d}, \text{exp}_\text{d})\}$ controlled by the head pose and facial expression from the driving sequence, while preserving the identity and background consistent with the source image.
Different from previous arts \cite{xnemo,fantasyportrait}, we also aim to disentangle the driving pose and expression to enable pose-only or expression-only editing as well as disentangled animation.

As illustrated in Fig.~\ref{fig:pipeline}, our method contains two steps:
\begin{enumerate}
    \item \textbf{Disentangled Motion Trainer.} 
    We design a powerful GAN-based motion trainer and augmentation strategies to learn disentangled pose and expression encoders.
    Specifically, we first transform and augment the driving image into the pose and expression image and extract the pose transformation and expression latent, respectively.
    A 3D appearance feature is encoded from the source image and warped from the source pose to driving pose.
    Then we modulate the warped feature by the expression latent through AdaIN \cite{huang2017adain}, producing an animated image and comparing it with the ground truth during training.
    \item \textbf{Diffusion-based Disentangled Animation.}
    We employ the latent diffusion model (LDM) \cite{SD1_5} as the backbone and injects the source identity through a reference UNet following \cite{hu2024animate, xnemo, champ}.
    Given the driving pose and expression signal, we propose a dual-branch pose injection with a cross-attention expression injection for disentangled portrait animation.
\end{enumerate}

\subsection{Preliminaries}\label{subsec:Preliminaries}
\noindent\textbf{Latent Diffusion Model (LDM).} LDM \cite{SD1_5} is a series of diffusion models \cite{ho2020ddpm,song2020denoising} that generate images in the latent space of pre-trained variational autoencoders (VAE).
During training, LDM corrupts a clean latent $z$ at time step $t$ with a Gaussian noise $\epsilon_t$ to obtain a noisy latent $z_t$, following DDPM \cite{ho2020ddpm}. 
Then a UNet-based \cite{ronneberger2015unet} denoising network $\hat\epsilon_\theta$ is then trained to predict $\epsilon_t$ under the condition $\bm c$ using an MSE loss:
\begin{equation}
    L_{\theta} = \mathbb E_{\epsilon\sim\mathcal N(0,1),t}\left[\|\epsilon_t-\hat\epsilon_\theta(z_t, \bm c;t)\|_2^2\right].
\end{equation}
At inference time, the model begins from a pure Gaussian noise and applies a multi‑step denoising process to generate meaningful latent samples according to the conditions.






\noindent\textbf{Reference UNet Architecture for Animation.} 
Owing to its powerful generative capabilities, LDM has been extensively adopted as the backbone network for motion-driven human animation synthesized from a single source image.
For skeletal animation, Animate Anyone \cite{hu2024animate} pioneered to introduce a reference UNet to extract fine-grained features from the source image, which are then injected into the denoising network through spatial attention.
Moreover, a temporal attention module \cite{guo2023animatediff} is also incorporated for temporal coherence. 
This paradigm was subsequently extended to portrait animation \cite{aniportrait, xportrait, xnemo}, a direction that our work also embraces.

\subsection{Disentangled Motion Trainer}
\label{subsec:Motion Trainer}

As shown in \cref{fig:pipeline} (a), we design a powerful GAN-based motion trainer and several augmentation strategies to learn precise and disentangled pose and expression encoders. 
This GAN consists of three encoders (including a 3D appearance encoder, an explicit pose encoder, and a latent expression encoder) and a StyleGAN2-like \cite{stylegan2} decoder.

\noindent\textbf{3D Appearance Encoder.}
Following \cite{facevid2vid,liveportrait, emoportraits}, a 3D feature containing the source identity is obtained through an appearance encoder instantiated by 2D and 3D CNNs.

\noindent\textbf{Explicit Pose Encoder.} 
We represent the head pose as an explicit global transformation $\mathbf{P}\in\mathbb{R}^{3\times 4}$ including rotation $\mathbf{R}$, translation $\mathbf{t}$ and scale $s$ (RTS) following \cite{facevid2vid,liveportrait}:
\begin{equation}
    \mathbf{P} = \begin{bmatrix}
s\mathbf{R} & \mathbf{t} 
\end{bmatrix}.
\label{eq:pose}
\end{equation}
From the perspective of representation, the transformation $\mathbf{P}$ with 6 ($3+2+1$) degrees of freedom refrains from the expression leakage.
Then we train a pose encoder instantiated by a ConvNeXt \cite{convnext} to extract RTS from the portrait images.
Obtaining the source and driving pose transformations ($\mathbf{P}_\text{s}$ and $\mathbf{P}_\text{d}$), we warp the 3D source feature from the source to driving pose under the transformation of $\mathbf{P}_\text{d}\mathbf{P}_\text{s}^{-1}$.


\noindent\textbf{Latent Expression Encoder.} 
We represent the facial expression as a 1D latent code with a dimension of 512 because of its expressive performance as demonstrated in X-NeMo \cite{xnemo}.
We extract the latent code from the input portrait image using an expression encoder instantiated by a face alignment network (FAN) \cite{bulat2017far}. 
Following the information bottleneck principle from previous works \cite{xnemo,pdfgc}, the expression latent refrains from the identity leakage. 
However, it may retain the head pose information leaked from the input image.

\begin{figure}
    \centering
    \includegraphics[width=0.48\textwidth]{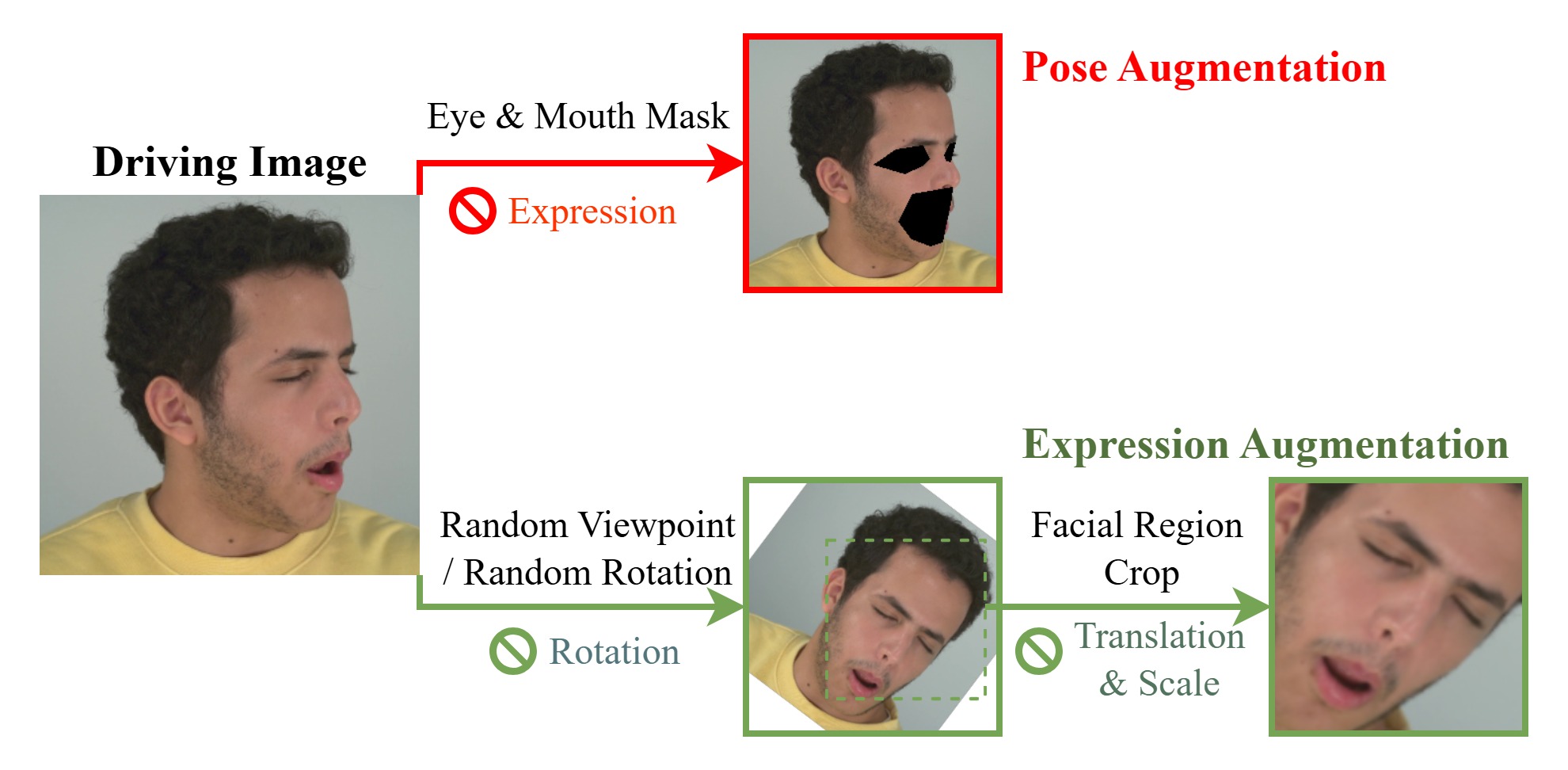}
    \caption{Illustration of the pose and expression augmentation.}
    \label{fig:augment}
\end{figure}

\noindent\textbf{Pose \& Expression Augmentation.}
To this end, we propose a series of augmentation strategies on the pose and expression driving image to prevent expression or pose leakage, as shown in \cref{fig:augment}.
For the pose input, we cover the eye and mouth regions using MediaPipe \cite{lugaresi2019mediapipe} landmarks to eliminate most expression information.
For the expression input, we first perform random rotation or select another viewpoint (applicable only to multi-view datasets), enabling the expression encoder to be insensitive to head rotation.
Then we crop the facial region and resize it to a fixed size of $224\times 224$ via the MediaPipe bounding box to eliminate the head translation and scale.
Overall, the above augmentation strategies promote the disentanglement between the pose and expression encoders.

\noindent\textbf{Expression-conditioned Decoder.} 
In the subsequent stage, we use a StyleGAN2-like generator to decode the warped appearance features into a target image while injecting the expression latent via AdaIN \cite{huang2017adain}. 

\subsection{Diffusion-based Disentangled Animation}\label{subsec:Diffusion}


After obtaining disentangled expression and pose encoders from the motion trainer, we employ a latent diffusion model (LDM) and a reference UNet architecture for portrait animation, as mentioned in Sec.~\ref{subsec:Preliminaries}.
Next, we elaborate on the methodology for injecting driving pose and expression signals into LDM to achieve disentangled and expressive animation.
Correspondingly, the overall framework is illustrated in \cref{fig:pipeline} (b).

\noindent\textbf{Dual-branch Pose Injection.}
Regarding the injection method for pose information, existing practices typically render the pose into 2D skeleton maps \cite{wananimate} or spheres \cite{luo2025dreamactor-m1} and then inject them into the denoising network in a spatial manner.
However, neither 2D skeletons nor spheres can accurately characterize the head pose.
Therefore, we propose two novel methods for pose injection: ray map and reference warping.

In the first branch, inspired by Pl\"ucker ray map \cite{plucker} that is widely used in camera pose controlled video generation \cite{he2024cameractrl,liang2025wonderland}, we propose to transform the head pose into a ray map, and concatenate the source and driving ray maps with the noisy latent for pose conditioned generation.
Given a head pose $\mathbf{P}$ (Eq.~\ref{eq:pose}), the ray map is formulated as
\begin{equation}
\begin{aligned}
    RayMap(u,v)&=\mathbf{P}[u,v,0,0]^\top - [u,v,0,0]^\top\\
    &(u,v)\in\left[-1,1\right]^2,
\end{aligned}
\end{equation}
\begin{figure}
    \centering
    \includegraphics[width=0.45\textwidth]{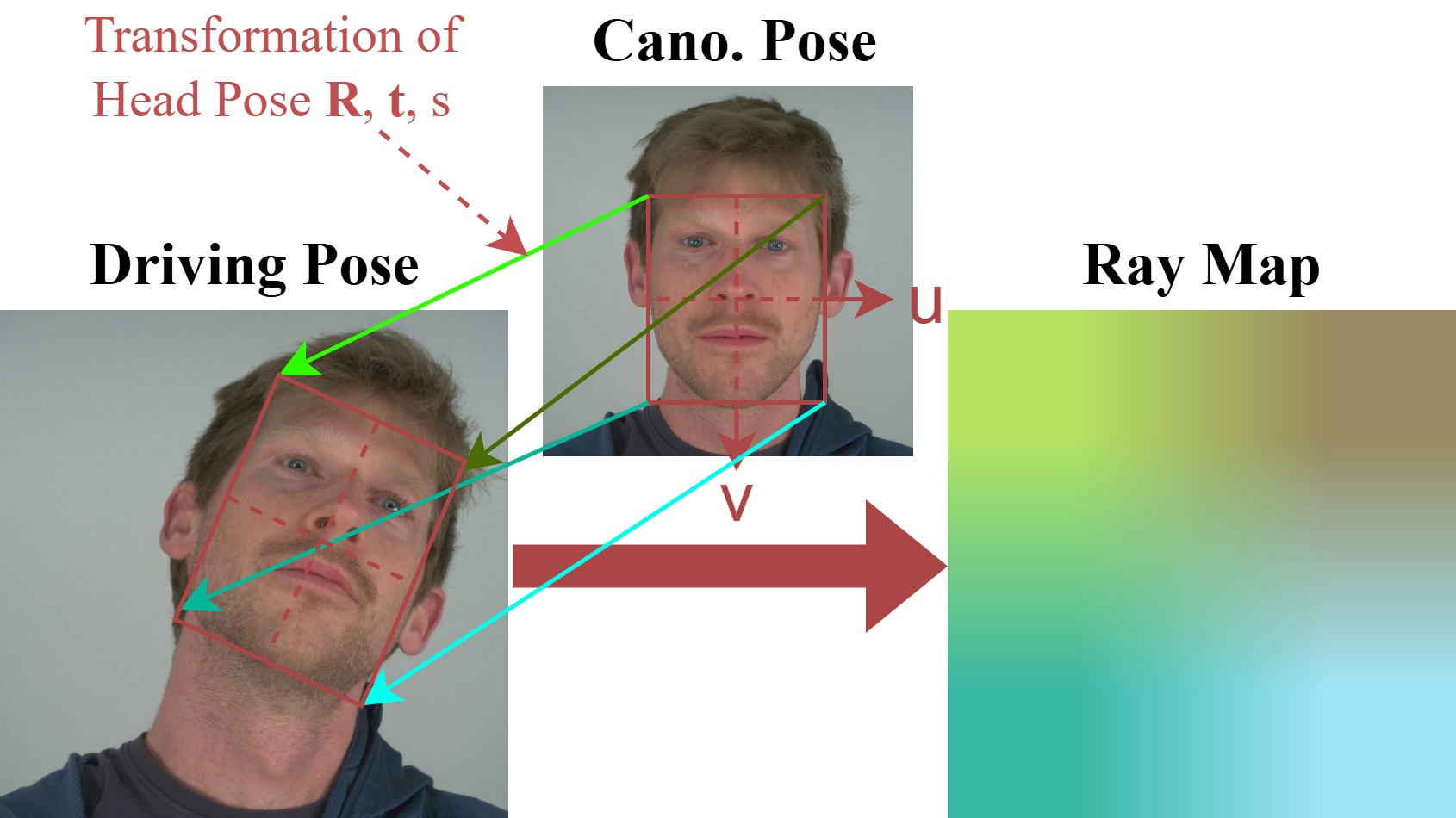}
    \caption{Illustration of the ray map of head pose.}
    \label{fig:raymap}
\end{figure}
where $(u,v)$ is the coordinate of each pixel in the ray map.
In terms of physical meaning, as illustrated in \cref{fig:raymap}, each pixel on the ray map represents a vector from the canonical head pose to a target one.
Given the spatial ray maps derived from both the source and driving poses, our method can achieve precise control over head rotation, translation, and scale, while guaranteeing identity consistency particularly in scenarios involving long-distance pose transitions.

However, we empirically found that relying solely on the ray map lead to edge misalignment between the synthesized result and the original image in the expression-only editing scenario, as illustrated in Fig.~\ref{fig:ablation on reference warping}.
We observe that LDM inherently possesses 3D perceptual capabilities, thereby allowing us to leverage pose signals for the direct manipulation of its intermediate latent features.
Specifically, we first reshape the 2D source features in the reference UNet into 3D tensors, then warp them from the source pose to the driving pose, and finally convert them back to 2D via a flatten operation.
Given that the warped source features are spatially aligned with the latent features in the denoising UNet, we first apply a convolutional projection layer to these warped features, then directly perform element-wise addition of the processed source features to the latent features in the denoising UNet.
By virtue of the reference warping-based injection mechanism, our method achieves more precise pose control, particularly in the expression-only editing scenario, as illustrated in Fig.~\ref{fig:ablation on reference warping}.

\noindent\textbf{Cross-attention Expression Injection.}
Since the expression latent is a global feature, we perform cross attention \cite{vaswani2017attention} between the latent features in the denoising UNet and the expression latent following \cite{xnemo,kirschstein2024diffusionavatars,kirschstein2025avat3r}.


\subsection{Progressive Hybrid CFG}

\begin{figure}
    \centering
    \includegraphics[width=1\linewidth]{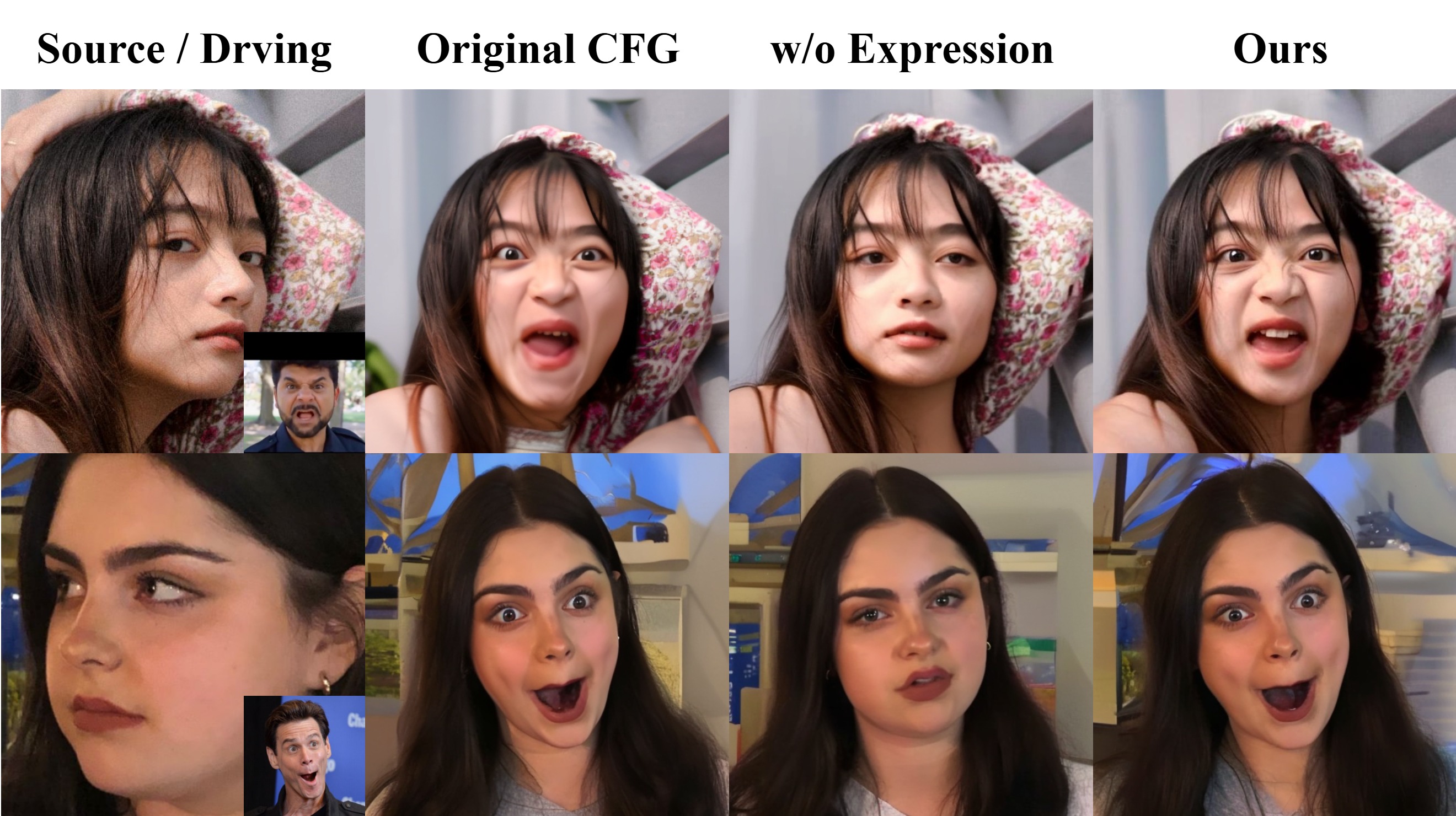}
    \caption{Pose-only generation preserves strong identity consistency (second from right). Compared with the original CFG, our method achieves better consistency with the source portrait (e.g., facial shapes) in scenarios involving significant pose and expression variations.}
    \label{fig:ablation on cfg}
\end{figure}

\begin{figure*}[t]
    \centering
    \includegraphics[width=1\textwidth]{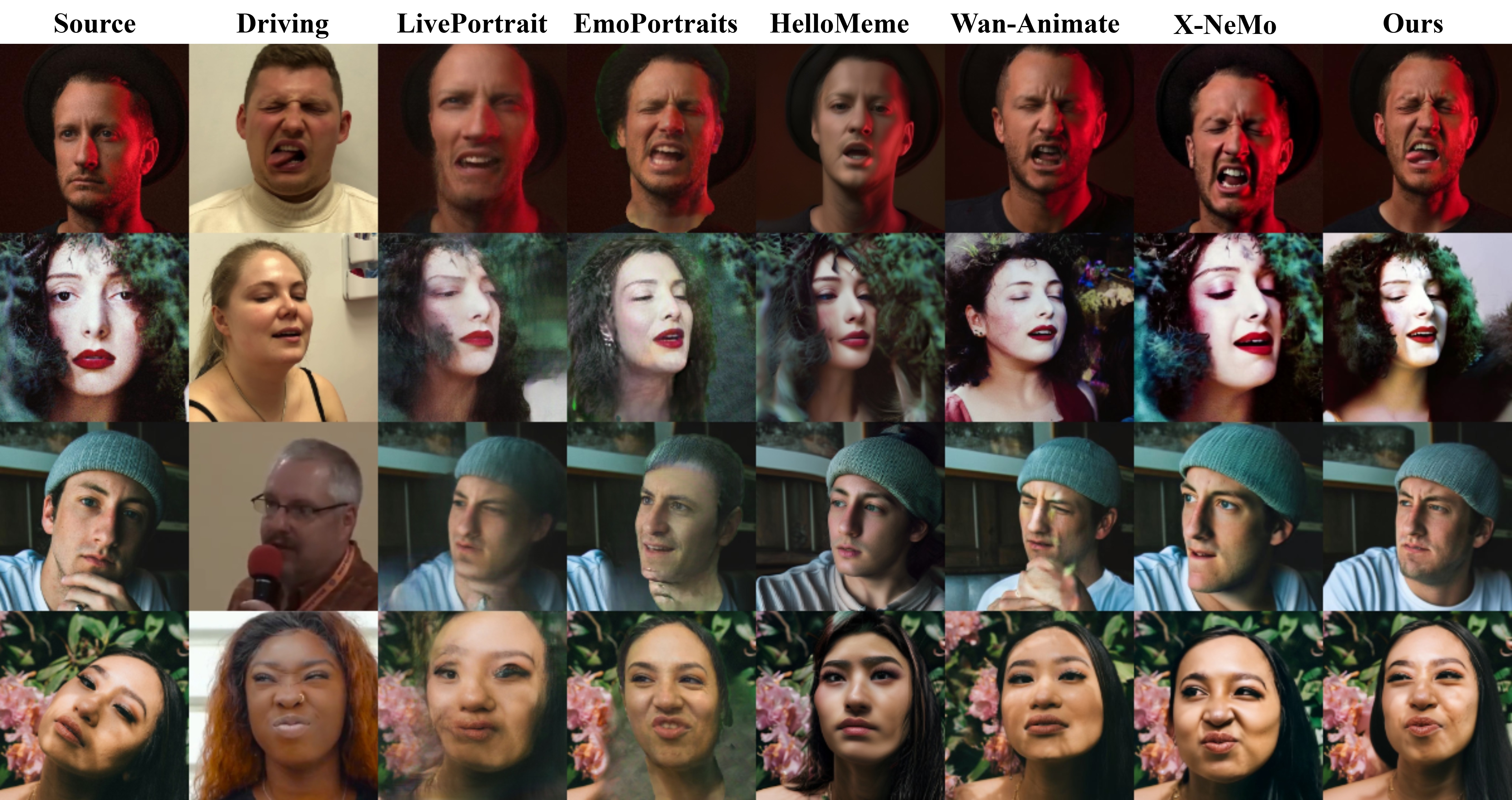}
    \caption{Qualitative comparison on cross-reenactment.}
    \label{fig:compare}
\end{figure*}

To enhance controllability and sample fidelity, classifier-free guidance (CFG) \cite{ho2022classifier} is commonly employed in the conditional diffusion process.
In the CFG scheme, each denoising step computes both a conditional noise estimation $\hat\epsilon_\theta(z_t, \bm c;t)$ and an unconditional one $\hat\epsilon_\theta(z_t,\emptyset;t)$. 
The final noise estimation is computed as
\begin{equation}
    \widetilde\epsilon_\theta(z_t, \bm c; t) \triangleq \omega\hat\epsilon_\theta(z_t, \bm c;t)+(1-\omega)\hat\epsilon_\theta(z_t, \emptyset;t),
\end{equation}
where $\omega=2.5$ denotes the CFG scale, a hyperparameter that regulates the conditioning strength.

We empirically found that the original CFG may yield unexpected results with inconsistency identity, especially when the source portrait faces sideways.
We hypothesize the underlying reason is that the identity, pose, and expression conditions are entangled at each step throughout the denoising process.
Drawing inspiration from \cite{ki2025float, Brooks_2023_CVPR}, we propose a progressive hybrid CFG that gradually incorporates pose and expression conditions.
Specifically, over the 35 steps of DDIM \cite{ddim}, we exclude the expression condition within the initial 5 steps; subsequently, we incorporate the expression condition incrementally across the next 5 steps; finally, we employ the full conditions for the remaining 25 steps:
\begin{equation}
\begin{aligned}
    &\widetilde\epsilon_\theta^\text{*}(z_t, \bm c; t) \triangleq \\
    &\begin{cases} 
     \widetilde\epsilon_\theta(z_t, \bm c|_\text{exp}; t) & 30\textless t\leq 35, \\
     \widetilde\epsilon_\theta(z_t, \bm c|_\text{exp}; t)\frac{t-25}{5} +
     \widetilde\epsilon_\theta(z_t, \bm c; t)\frac{30-t}{5} &25\textless t\leq30\\
     \widetilde\epsilon_\theta(z_t, \bm c; t) &t\leq 25,
    \end{cases},
\end{aligned}
\end{equation}
where $\bm c$ denotes all the conditions including identity, head pose, expression, $\bm c|_\text{exp}$ represents the conditions excluding expression, and $\widetilde\epsilon_\theta^\text{*}$ is our final noise estimation. 
As illustrated in Fig.~\ref{fig:ablation on cfg}, our progressive CFG can produce expressive animation with more consistent identity.

\begin{figure*}
    \centering
    \includegraphics[width=0.875\textwidth]{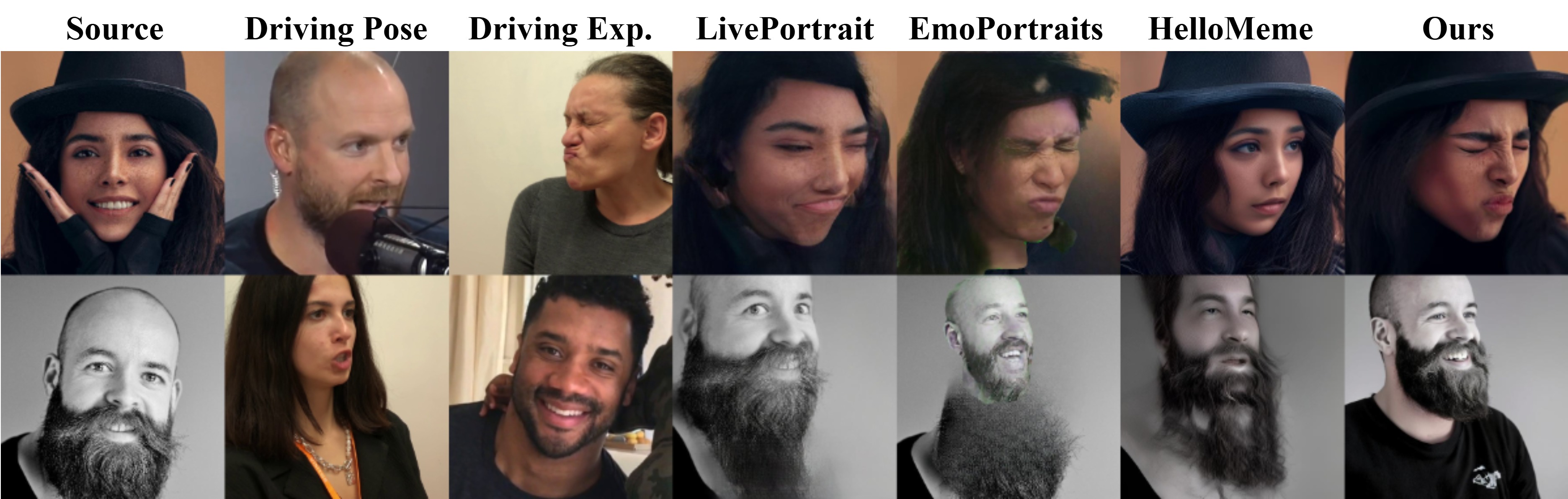}
    \caption{Qualitative comparison on disentangled-reenactment.}
    \label{fig:compare_dc}
\end{figure*}

\begin{table*}[htbp]
\centering
\begin{tabular}{@{}lccccccccc@{}}
\toprule
\multirow{2}{*}{Method} & \multicolumn{3}{c}{Self-Reenactment}              & \multicolumn{3}{c}{Cross-Reenactment}             & \multicolumn{3}{c}{Disentangled-Reenactment}        \\ \cmidrule(l){2-10} 
                        & PSNR↑          & SSIM↑          & LPIPS↓          & CSIM↑          & AED↓            & APD↓           & CSIM↑          & AED↓            & APD↓           \\ \midrule
EMOPortraits \cite{emoportraits}            & 20.010         & 0.758          & 0.248           & 0.255          & 0.0561          & 0.264          & 0.217          & \cellcolor{orange!20}{0.0586}    & \cellcolor{orange!20}{0.155}    \\
LivePortrait \cite{liveportrait}            & 27.760         & 0.857          & \cellcolor{orange!20}0.098           & 0.459          & 0.0696          & 0.236          & \cellcolor{orange!20}{0.458}    & 0.0695          & 0.195          \\
HelloMeme \cite{hellomeme}              & 18.870         & 0.660          & 0.292           & 0.313          & 0.0772          & \cellcolor{orange!20}0.173    & 0.302          & 0.0874          & 0.226          \\
X-NeMo \cite{xnemo}                 & 21.060         & 0.741          & 0.207           & 0.492          & \cellcolor{orange!20}{ 0.0518}    & 0.551          & N/A              & N/A               & N/A              \\
Wan-Animate \cite{wananimate}            & \cellcolor{orange!20}{27.970}   & \cellcolor{orange!70}{0.865} & \cellcolor{orange!20}{0.098}     & \cellcolor{orange!20}{ 0.551}    & 0.0588          & 0.180          & N/A             & N/A               & N/A              \\
Ours                    & \cellcolor{orange!70}{28.590} & \cellcolor{orange!20}{0.862}    & \cellcolor{orange!70}{0.088} & \cellcolor{orange!70}{0.623} & \cellcolor{orange!70}{0.0515} & \cellcolor{orange!70}{0.145} & \cellcolor{orange!70}{0.631} & \cellcolor{orange!70}{0.0546} & \cellcolor{orange!70}{0.100} \\ \bottomrule
\end{tabular}
\caption{Quantitative comparisons between our method and baselines. ``N/A'' means X-NeMo and Wan-Animate do not support disentangled reenactment. We highlight the best scores with orange shading, and the second best with light orange.}
\label{tab:benchmarks}
\end{table*}

\section{Experiment}
\subsection{Implenmentation details}
We utilize two multi-view portrait video datasets (NerSemble \cite{Kirschstein_2023}, ava-256 \cite{martinez2024codec}) and two in-the-wild monocular datasets (PFHQ \cite{chen2023realworld}, VFHQ \cite{wang2022vfhqhighqualitydatasetbenchmark}) for joint training. 
We process them to a fixed resolution of 512 \( \times \) 512. 

The whole training process involves 3 stages: 
\begin{enumerate}
    \item Motion Training. This stage is trained to obtain precise and disentangled expression and pose encoders, with a batch size of 112 and a learning rate of $1\times 10^{-4}$ for 200k iterations. 
    \item Diffusion Training. During this phase, we freeze the expression and pose encoders and train the reference and denoising UNets with a batch size of 48 and a learning rate of $1\times 10^{-5}$ for 120k iterations.
    \item Temporal Training. Only the temporal module is trained using 24-frame video sequences, with a batch size of 8 and a learning rate of $1\times 10^{-5}$ for 80k iterations. 
\end{enumerate}

\subsection{Benchmark, Baselines and Metrics}
\noindent\textbf{Benchmark.}
We collected a total of 150 copyright-free in-the-wild portrait photos from Life of Pix \cite{PixofLife} and Unsplash \cite{UnSplash}, covering different ethnicities, lighting, poses and expressions. 
Meanwhile, we gathered 150 video clips (covering large-scale pose and expression variations) from a portrait video dataset TalkingHead1kH \cite{wang2021facevid2vid} and an extreme expression dataset FEED \cite{emoportraits}. 

\noindent\textbf{Baselines.}
We compare our method against state-of-the-art baselines, including diffusion-based methods (Wan-Animate \cite{wananimate}, X-NeMo \cite{xnemo} and HelloMeme \cite{hellomeme}) and GAN-based methods (LivePortrait \cite{liveportrait} and EMOPortraits \cite{emoportraits}). 
Among them, LivePortrait, EMOPortraits and HelloMeme can disentangle the pose and expression controls.

\noindent\textbf{Metrics.}
We adopt PSNR, SSIM \cite{wang2004ssim} and LPIPS \cite{zhang2018lpips} to evaluate the differences between the prediction and ground truth for self-reenactment scenarios.
Since there exists no ground truth for cross-reenactment and disentangled-reenactment, we utilize CSIM \cite{deng2019arcface}, Average Expression Distance (AED) and Average Pose Distance (APD) computed using MediaPipe \cite{lugaresi2019mediapipe} for evaluating identity similarity, expression and pose accuracy. 

\begin{figure}[t]
    \centering
    \includegraphics[width=0.48\textwidth]{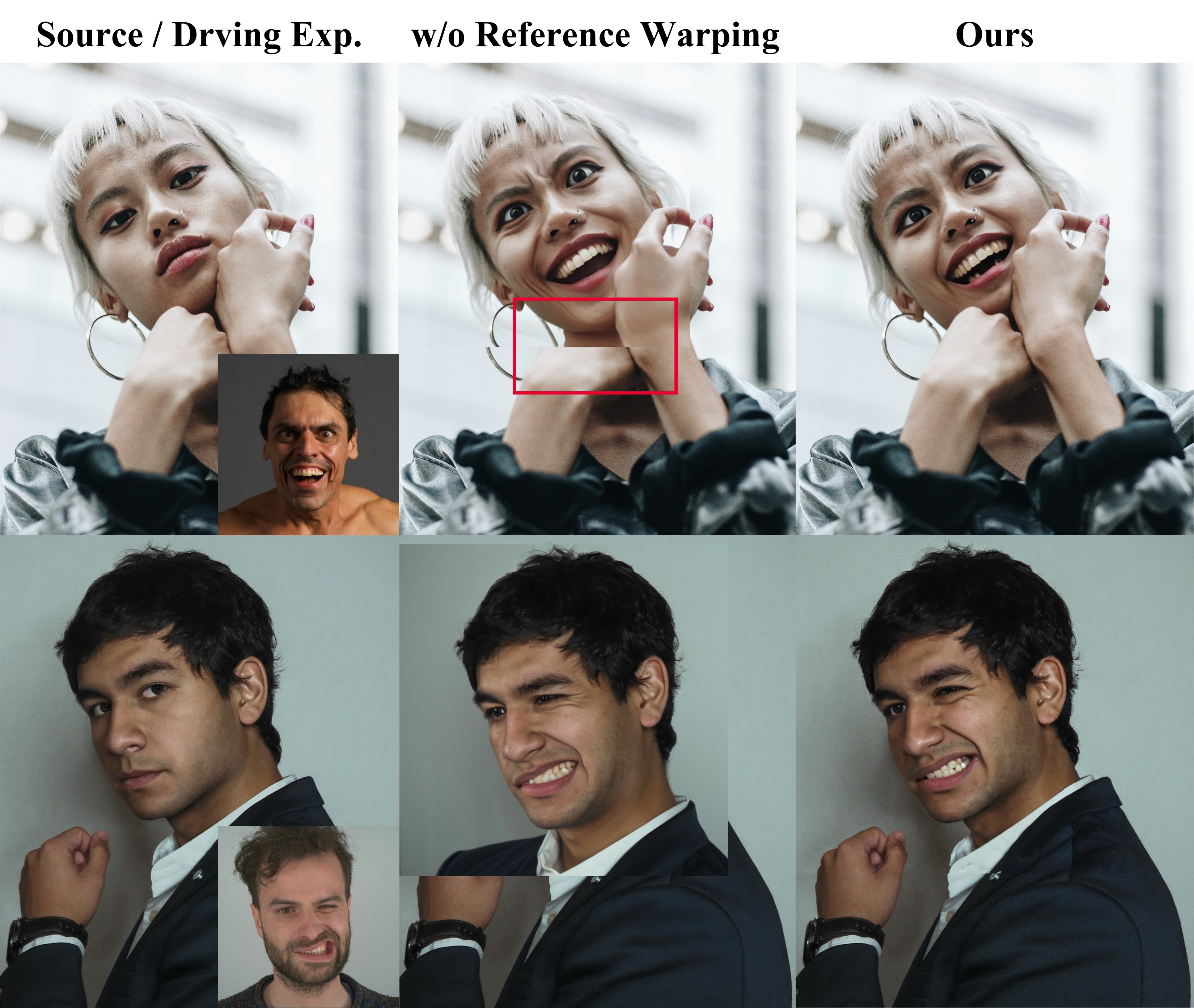}
    \caption{Qualitative ablation study of reference warping on the expression-only editing scenario.}
    \label{fig:ablation on reference warping}
\end{figure}

\subsection{Comparison}
\noindent\textbf{Self-Reenactment.}
In each collected video sequence, we select one frame as the reference image and use the other frames as the driving video to test the self-reenactment performance of different models. 
Then we reuse the driving video as the ground truth to compute PSNR, SSIM, and LPIPS.
\cref{tab:benchmarks} demonstrate that our method outperforms other approaches on PSNR and LPIPS, with only a marginal deficit in SSIM compared to Wan-Animate.

\noindent\textbf{Cross-Reenactment.}
We utilize portrait photos as source images and video clips as driving videos from our benchmark for cross-reenactment comparison. 
We present qualitative comparison in \cref{fig:compare}. 
In contrast to other methods, our approach excels in the identity consistency, pose accuracy (especially the scale and translation) and expressiveness (e.g., tongues and squinting).
The quantitative metrics are also reported in Tab.~\ref{tab:benchmarks}.
GAN-based methods including LivePortrait \cite{liveportrait} and EmoPortraits \cite{emoportraits} suffer from blurriness and motion distortion due to their limited generative capability.
Among the diffusion-based methods, HelloMeme \cite{hellomeme} struggles to capture nuanced facial motions, as it utilizes a CLIP-based \cite{clip} motion encoder that is unsuitable for the specific task of portrait animation.
Wan-Animate \cite{wananimate} leverages a pre-trained expression encoder from LIA \cite{lia} to enable expression control; however, its performance is constrained by a limited set of only 20 linear expression bases.
X-NeMo \cite{xnemo}, one of the state-of-the-art portrait animation baselines, delivers realistic and expressive animation, yet struggles to accurately reenact the driving pose, especially positional translation and scale variations, attributed to its motion representation that combines an entangled latent and a simplistic spatial triplet.
Overall, thanks to the explicit pose and latent expression representations and effective injection mechanism, our method realizes expressive and precise controls over head pose and facial expression for cross-reenactment scenarios.

\noindent\textbf{Disentangled-Reenactment.} 
To verify the disentangled controllability of our model, we compare our method with related works (including LivePortrait \cite{liveportrait}, EmoPortraits \cite{emoportraits}, and HelloMeme \cite{hellomeme}) that support disentangled animation capabilities, using distinct driving pose and expression inputs.
Given a source portrait photo $\mathbf{I}_\text{s}$, we select two different video clips from the bench mark as the driving pose $\{\mathbf{I}_\text{d}^\text{pose}\}$ and expression $\{\mathbf{I}_\text{d}^\text{exp}\}$, respectively, producing animation results $\{\hat{\mathbf{I}}\}$.
We evaluate identity consistency via CSIM between $\{\hat{\mathbf{I}}\}$ and $\mathbf{I}_\text{s}$, pose control accuracy via APD between $\{\hat{\mathbf{I}}\}$ and $\{\mathbf{I}_\text{d}^\text{pose}\}$, and expression control accuracy via AED between $\{\hat{\mathbf{I}}\}$ and $\{\mathbf{I}_\text{d}^\text{exp}\}$, respectively.
We present qualitative and quantitative comparisons in \cref{fig:compare_dc} and \cref{tab:benchmarks}, respectively. 
Both comparisons demonstrate that our method achieves precise control over pose and expression, while preserving identity consistency with the source portrait.
Moreover, thanks to the powerful disentangled capability, our method also enables expression-only and pose-only editing, as shown in \cref{fig:teaser} and the Supp. Mat.


\subsection{Ablation Study}
\begin{figure}
\centering  
    \includegraphics[width=0.8\linewidth]{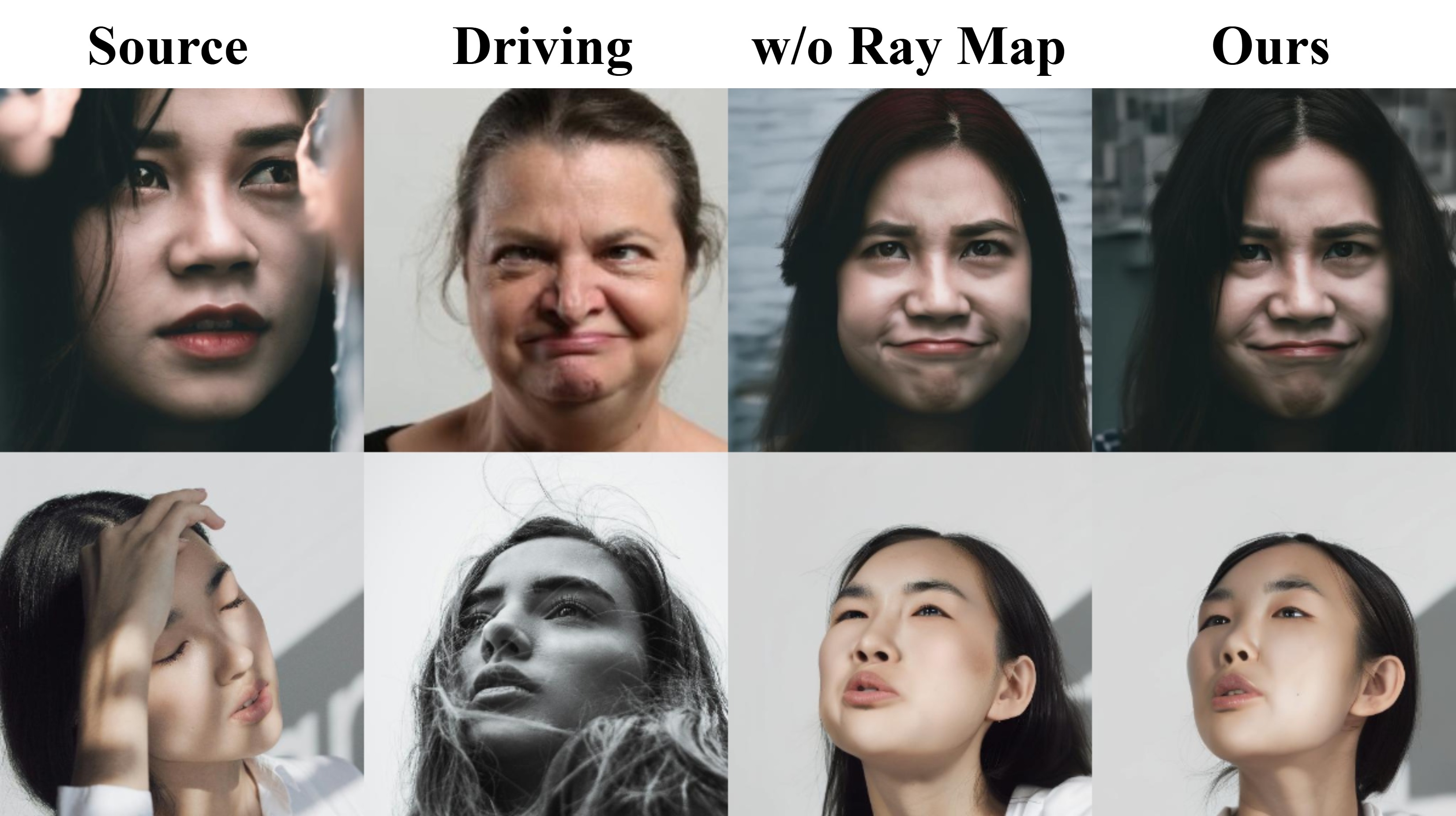}
    \caption{Qualitative ablation of the head pose ray map.}
    \label{fig:wo_raymap}
    \includegraphics[width=\linewidth]{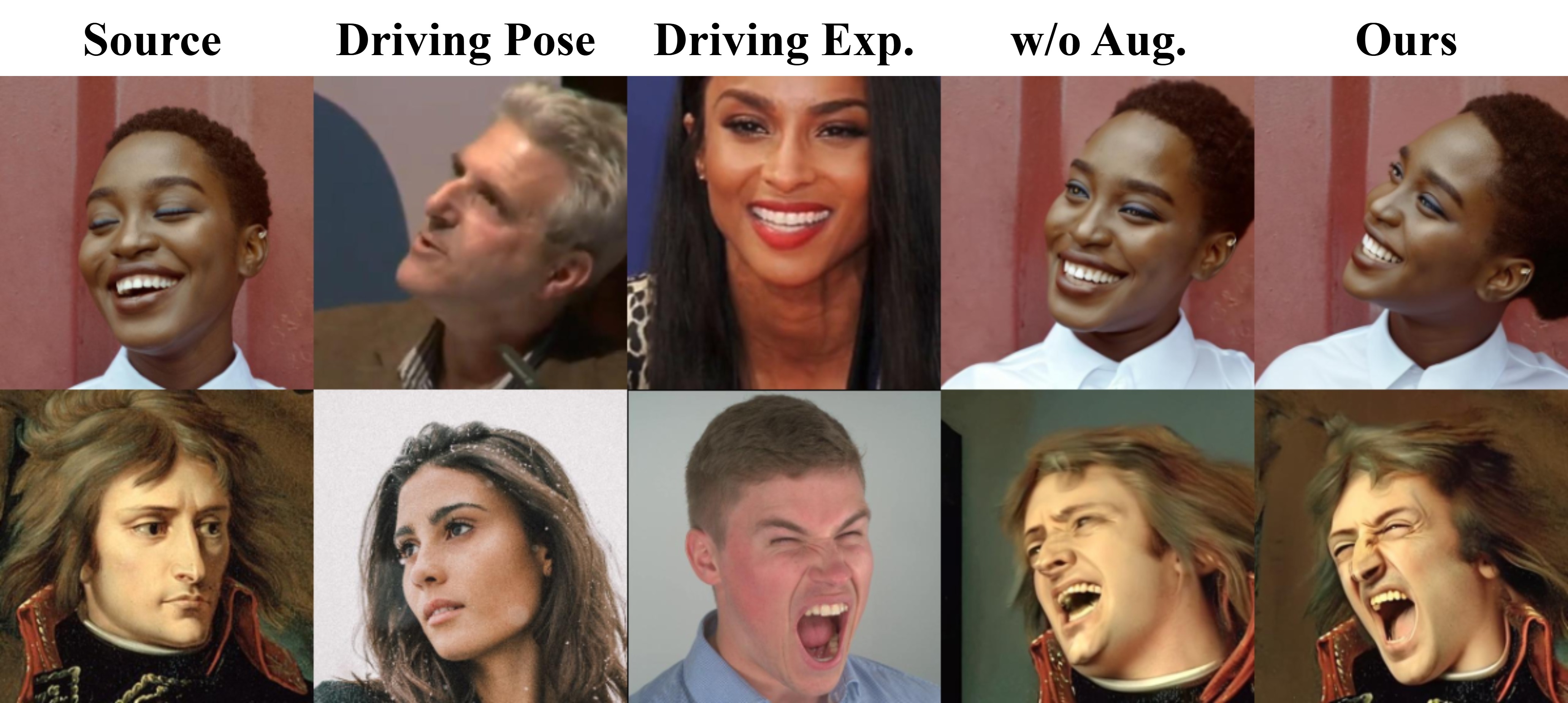}
    \caption{Qualitative ablation of the augmentations.}
    \label{fig:wo_aug}
\end{figure}

\noindent\textbf{Head Pose Ray Map.}
We evaluate the head pose ray map by eliminating it, i.e., using only the reference warping for pose injection.
Fig.~\ref{fig:wo_raymap} illustrates that the ray maps provide long-range correspondences between the source and driving poses, thereby enabling more stable identity consistency, especially when the source and driving rotations and scales exhibit substantial differences.

\noindent\textbf{Reference Warping.}
We found that relying solely on the ray map for pose injection could result in inconsistent boundaries and backgrounds in expression-only editing scenarios, which leads to visible seams when pasted back onto the original full image, as shown in Fig.~\ref{fig:ablation on reference warping}.
Reference warping delivers a robust signal of an identity matrix derived by the pose transformation in this scenario, thereby enabling expression-only modification.

\noindent\textbf{Pose \& Expression Augmentation.}
As shown in \cref{fig:wo_aug}, it can be observed that after removing the pose \& expression augmentation, the model’s generation consistency for expressions and poses exhibits a significant decrease. This indicates that the proposed strategy can effectively guide the pose and expression extractors to extract their corresponding features, thereby avoiding the accuracy loss caused by the mutual interference between pose information and expression information.

\noindent\textbf{Quantitative Ablation Studies.}
All the numerical results are reported in \cref{tab:ablation}.
It demonstrates that these core contributions yield the most consistent identity, the most accurate pose reenactment, and high-fidelity expressiveness.



\begin{table}[t]
\resizebox{0.48\textwidth}{!}{
\begin{tabular}{@{}lcccccc@{}}
\toprule
\multirow{2}{*}{Method} & \multicolumn{2}{c}{Cross-Reenactment}  & \multicolumn{2}{c}{Disentangled-Reenactment} \\ \cline{2-5} 
                        & CSIM↑          & AED/APD↓              & CSIM↑            & AED/APD↓                \\ \hline
w/o rap map            & 0.609          & \textbf{0.0506}/0.162 & 0.609            & \textbf{0.0542}/0.105    \\ 
w/o warping             & 0.619          & 0.0507/0.166          & 0.631            & 0.0573/0.121            \\
w/o augmentation            & 0.619          & 0.0583/0.283 & 0.629            & 0.0634/0.168    \\
Ours                    & \textbf{0.623} & 0.0515/\textbf{0.145} & \textbf{0.631}   & 0.0546/\textbf{0.100}     \\ \hline
\end{tabular}
}
\caption{Quantitative ablation studies.}
\label{tab:ablation}
\end{table}

\section{Discussion}
\noindent\textbf{Conclusion.}
We present DeX-Portrait, a new diffusion-based portrait animation framework that leverages explicit and latent motion representations for both disentangled and expressive animation.
We propose to represent head pose and facial expression as a global transformation and a latent code, respectively, and design a dedicated motion trainer along with augmentation strategies for learning mutually disentangled pose and expression encoders.
Based on the pretrained motion encoders, we propose a dual-branch pose injection mechanism coupled with cross-attention based expression injection—tailored for the diffusion model—enabling precise and independent control over portrait animation.
Overall, our method outperforms other state-of-the-art approaches on both expressiveness and disentangled controllability.

\noindent\textbf{Limitation.}
Our model is exclusively trained on real human datasets, and thus lacks generalization ability to other styles, e.g., cartoons.
Moreover, our model struggles to perform reliably in scenarios involving multiple portraits and significant occlusions.

\noindent\textbf{Ethics Statement.}
We acknowledge that our method could potentially be exploited to produce synthetic misinformation videos. Thus we emphasize the necessity of exercising responsible use of this technology, accompanied by clear synthetic content disclaimers.

{
    \newpage
    \small
    \bibliographystyle{ieeenat_fullname}
    \bibliography{main}
}

\clearpage
\appendix
\setcounter{page}{1}
\maketitlesupplementary

\renewcommand{\thefigure}{S\arabic{figure}}
\renewcommand{\thetable}{S\arabic{table}} 

\setcounter{figure}{0}
\setcounter{table}{0}

This document provides additional implementation details and experimental results. We include detailed implementation information for the Motion Trainer, Portrait Animator, dataset usage, and more results. We encourage the reader to examine our visual results for further insights.

\begin{figure*}[t]
    \centering
    \includegraphics[width=1\textwidth]{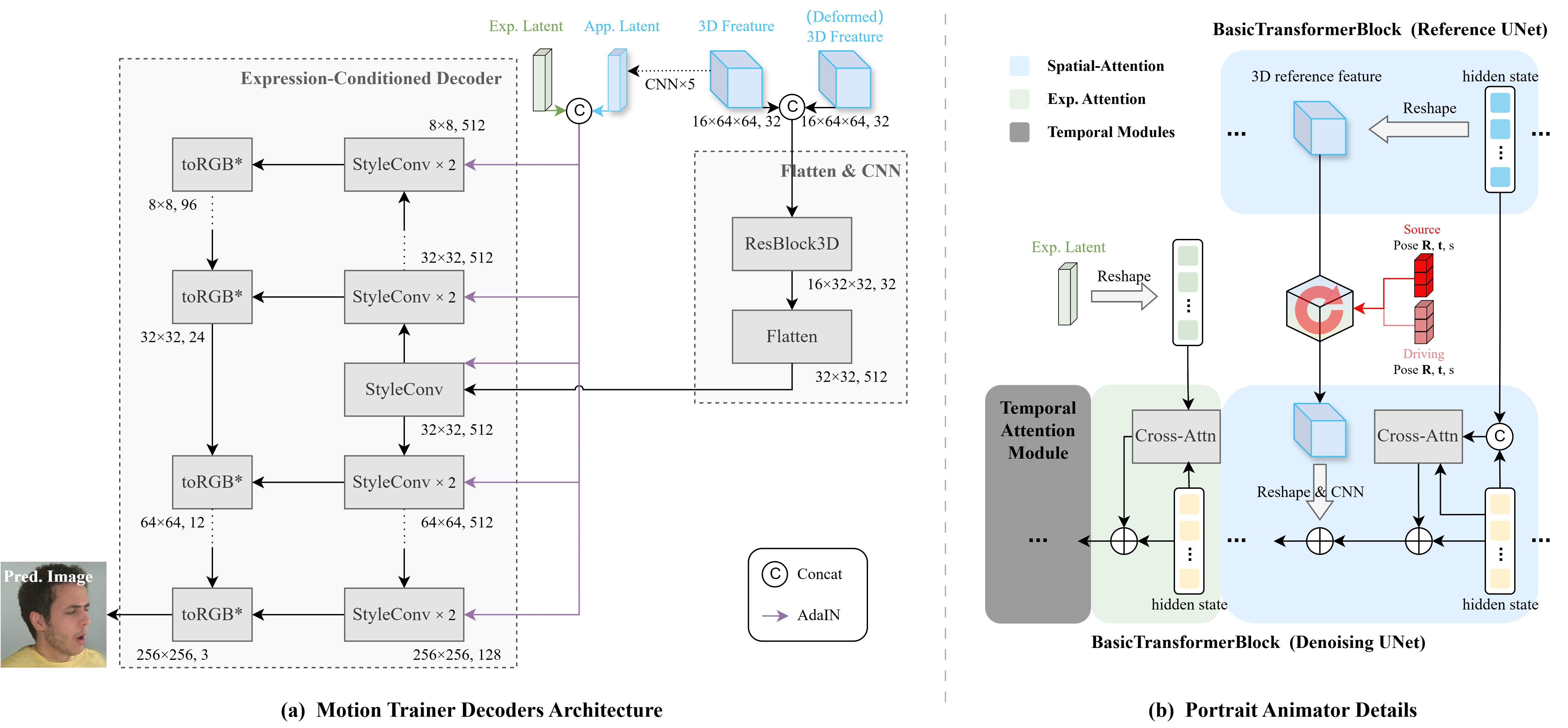}
    \caption{Detailed architecture of DeX-Portrait.}
    \label{fig:decoder}
\end{figure*}

\section{Implementation Details}
In this section, we elaborate on the architectures of Expression Conditioned Decoder and the Flatten \& CNN block in the Motion Trainer (Fig. 2 (a)) and provide implementation details in the Diffusion-Based Portrait Animator (Fig. 2 (b)).

\subsection{Motion Trainer Architecture}
The 3D features (Fig. 2 (a)) are processed through two sequential modules and ultimately decoded into a 256-resolution image, as shown in \cref{fig:decoder} (a). In the Flatten \& CNN block, we first concatenate the 3D feature and the warped 3D feature (both with spatial sizes 16 \( \times \) 64 \( \times \) 64) along the channel dimension. This is followed by a ResBlock3D module \cite{liveportrait} that compresses the spatial sizes to 16 \( \times \) 32 \( \times \) 32 while reducing the channel count from 64 to 32. Finally, the features are transformed into 2D representations via the flatten module.

Next, we decode these 2D appearance features into an image using the Expression-Conditioned Decoder. The decoder architecture introduces two key enhancements compared to StyleGAN2 \cite{stylegan2}: 
\begin{itemize}
    \item Intermediate Feature Injection: To maximize appearance preservation, the 2D features are injected into the intermediate layer (4th layer) of the decoder. The 3 \( \times \) 3 StyleConv modules simultaneously perform upsampling and downsampling to capture multi-scale details.
    \item Progressive Channel Reduction: To better blend multi-resolution images, we modify the toRGB module (*toRGB) to gradually reduce the channel count from 96 to 3 during the upsampling process from 8 \( \times \) 8 to 256 \( \times \) 256.
\end{itemize}
Additionally, 3D features are first processed through a 5-layer CNN, then flattened into a 1D 2048-dimensional appearance latent code, which is concatenated with a 1D 512-dimensional expression latent code obtained from the expression encoder. The combined vector serves as a style code for subsequent StyleConv modules.

\subsection{Portrait Animator Implementation Details} 
In the Portrait Animator, both our reference UNet and denoising UNet are based on the SD1.5 \cite{sd2022} architecture. 
We now detail the injection mechanisms for reference, head pose, and expression information in our implementation.
Head pose and reference information are primarily injected through the spatial-attention module, as illustrated in the \cref{fig:decoder} (b). Unlike Animate Any One \cite{hu2024animate}, which relies solely on cross-attention for injection, we further incorporate a residual connection to inject the warped appearance features.
Specifically, we modify each BasicTransformerBlock in SD1.5. In the reference unet, the hidden state has a token count of $h^2$ and channel dimension $c$, which is reshaped into a 3D appearance feature of size $\frac h2\times h\times h$ with $\frac{2c}h$ channels. 
This feature is then warped using the source and target head pose $\mathbf R, \mathbf t, s$, passed through a convolutional layer, and finally injected into the denoising UNet via a residual connection.
For expression latent injection, we follow the implementation of X-Nemo \cite{xnemo} by reshaping the 512-dimensional 1D expression latent into 32 tokens of 16-dimensional embeddings. These tokens are then injected into the denoising UNet through an additional cross-attention mechanism. A temporal attention module is further introduced at the final stage to enhance temporal coherence.

\section{Training Strategies}
\subsection{Motion Trainer Losses}
We train our motion trainer in a self-supervised manner on video datasets. Specifically, we select a source image $\mathbf{I}_\text{s}$ and a driving image $\mathbf{I}_\text{d}$ (serving as ground truth) from the video dataset and supervise the predicted image $\hat{\mathbf{I}}$ using a combination of multiple losses:
\begin{itemize}
    \item Reconstruction Loss. We minimize pixel-wise color differences with:
    \begin{equation}
        \mathcal L_1 = \|\mathbf{I}_\text{d} - \hat{\mathbf{I}}\|_1
    \end{equation}
    \item VGG16-based LPIPS Loss \cite{simonyan2014very}. To enhance perceptual realism and sharpness, we introduce:
    \begin{equation}
        \mathcal L_{lpips} = \sum_{i=1}^N\|VGG^i(\mathbf{I}_\text{d}) - VGG^i(\hat{\mathbf{I}})\|^2
    \end{equation}
    where $N$ denotes the number of feature layers in each respective pre-trained VGG model.
    \item Component LPIPS Loss for Facial Regions. A face-aware loss focusing on facial details:
    \begin{equation}
        \mathcal L_{clpips} = \sum_{i=1}^N\bm \|M\odot(VGG^i(\mathbf{I}_\text{d}) - VGG^i(\hat{\mathbf{I}}))\|^2
    \end{equation}
    where $\bm M$ is the facial segmentation mask.
    \item StyleGAN2-based Discriminator Loss. We employ a co-trained StyleGAN2 discriminator $D$ to calculate adversarial loss:
    \begin{equation}
        \mathcal L_{adv} = Softplus(-D(\hat{\mathbf{I}}))
    \end{equation}
\end{itemize}
Finally, the total loss is combined as:
\begin{equation}
    \mathcal L_{total} = \lambda_r\mathcal L_1 + \lambda_{lpips}\mathcal L_{lpips} + \lambda_{clpips}\mathcal L_{clpips} + \mathcal L_{adv}
\end{equation}
where $\lambda_r=10$, $\lambda_{lpips}=1$, $\lambda_{clpips}=100$.
\subsection{Dataset Utilization}
During the Motion Training and the Diffusion Training stages, beyond the previously mentioned pose and expression augmentation strategies, we further employ cross-view training for the multi-view datasets NeRSemble and ava-256. Specifically, within each training batch, one source frame is paired with four driving frames, which may collected from different camera viewpoints of the same subject. This training paradigm artificially simulates extreme head pose variations, thereby expanding the distribution range of our training data. For temporal module training, each batch consists of one source frame and 24 consecutive driving frames, with video sequences containing significant background motion being dynamically filtered out through real-time computation.

\section{More Results}

\begin{figure*}[t]
    \centering
    \includegraphics[width=1\textwidth]{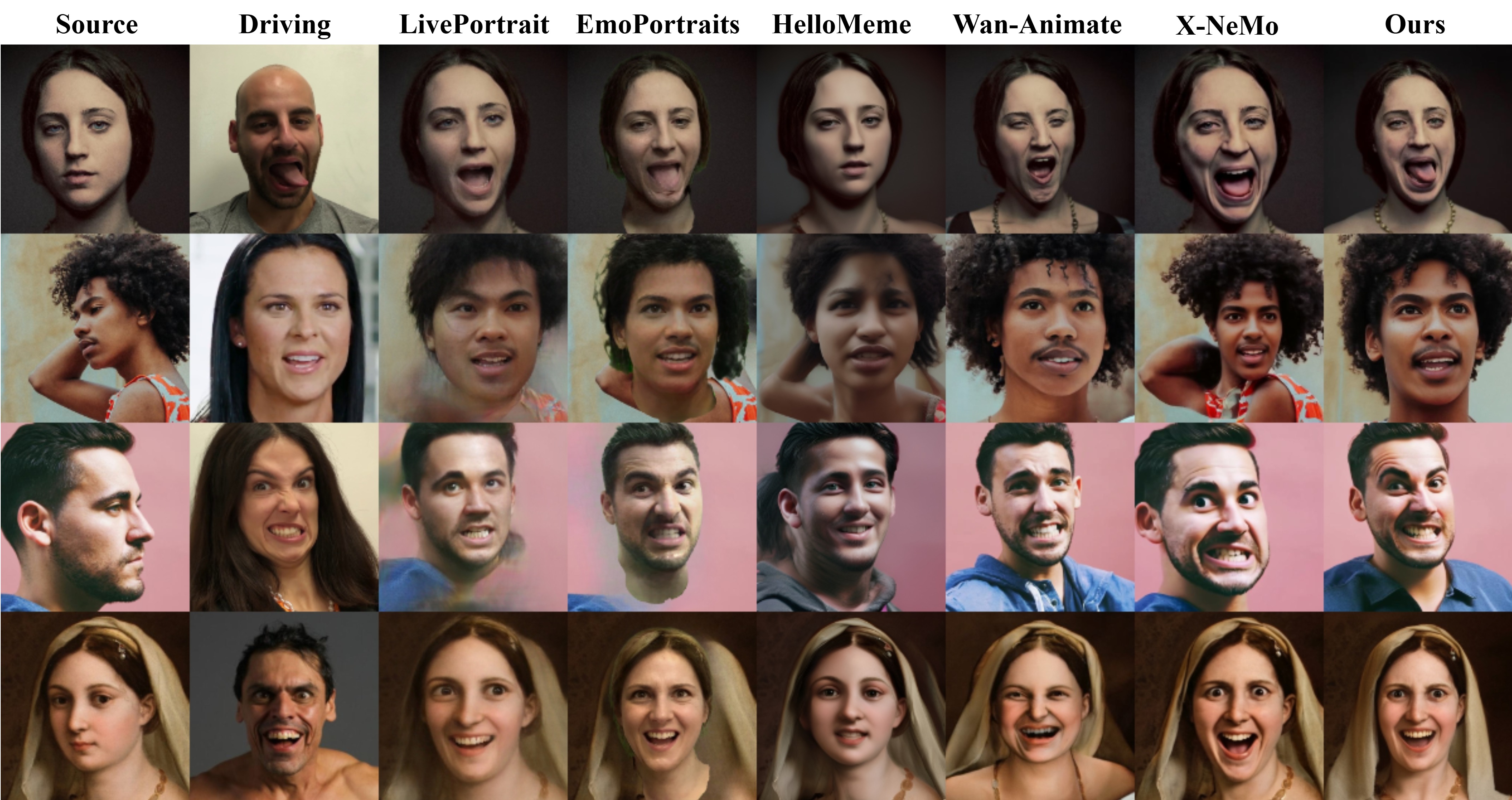}
    \caption{Qualitative comparison on cross-reenactment.}
    \label{fig:supp_compare}
\end{figure*}

For the cross-reenactment scenario, we additionally present comprehensive comparisons against state-of-the-art (SOTA) methods on benchmark datasets to demonstrate the superiority of our framework in identity preservation and precision control of head pose and expression \cref{fig:supp_compare}.

\begin{figure}[t]
    \centering
    \includegraphics[width=1\linewidth]{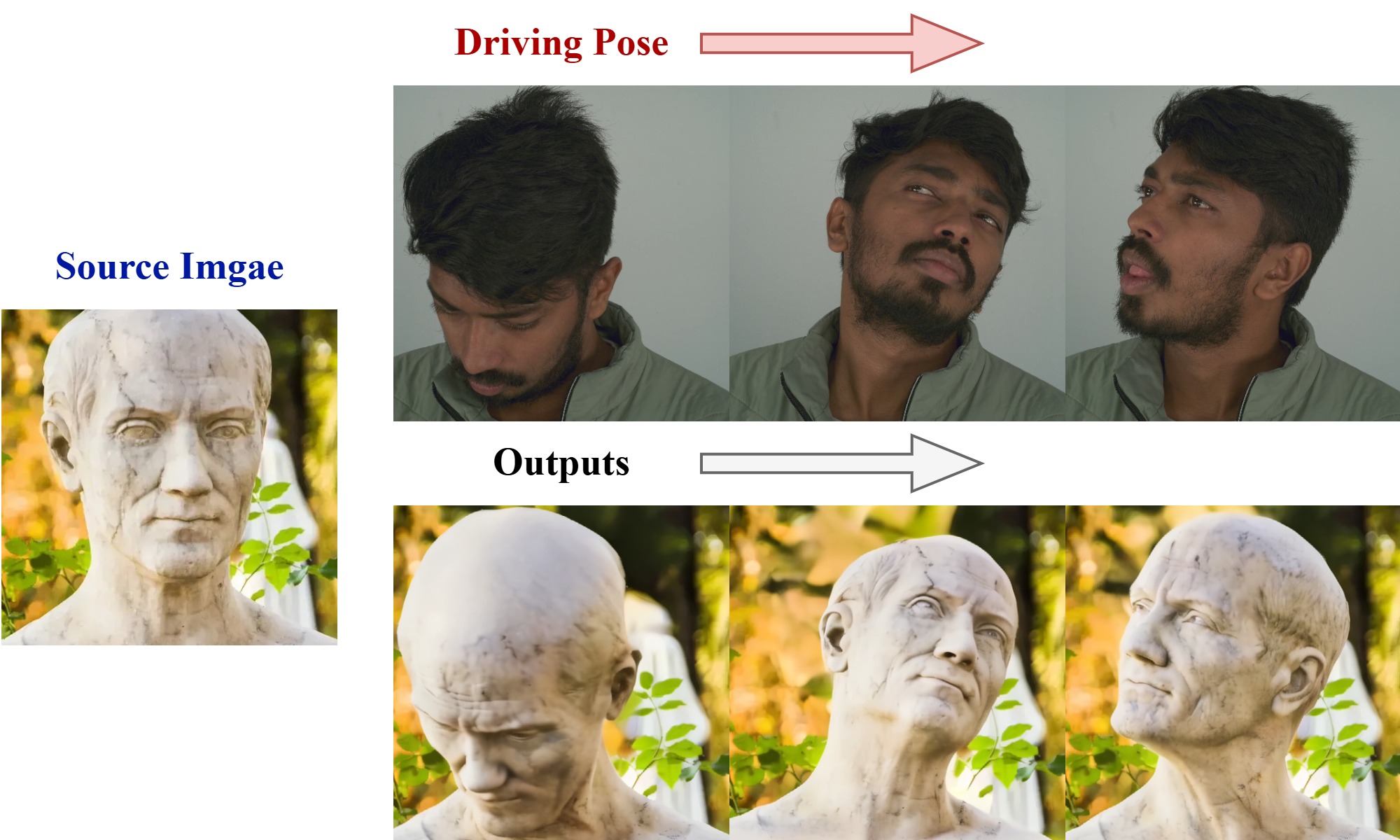}
    \caption{Visual results of pose-only editing.}
    \label{fig:supp_poe}
\end{figure}
\begin{figure}[t]
    \centering
    \includegraphics[width=1\linewidth]{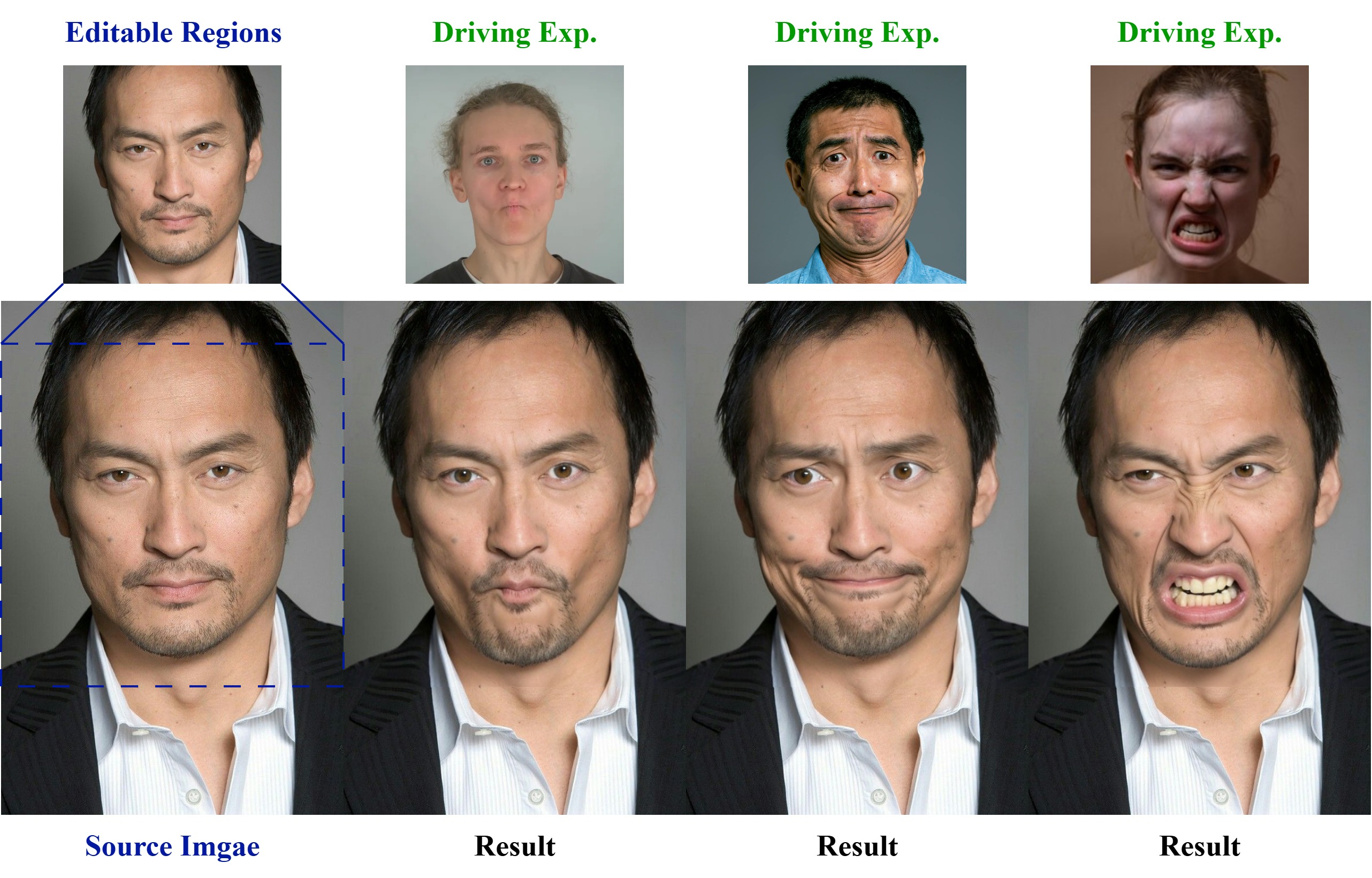}
    \caption{Visual results of expression-only editing.}
    \label{fig:supp_eoe}
\end{figure}

We also present visual demonstrations of two key applications enabled by our framework: expression-only editing and pose-only editing. As illustrated in \cref{fig:supp_eoe}, expression-only editing selectively modifies pixel values within local facial regions surrounding the target face. Owing to the high precision of our framework, the edited content can be seamlessly paste back into the original facial area without visible artifacts. \cref{fig:supp_poe} demonstrates pose-only editing capabilities, which enable users to reorient head poses in portrait images while strictly preserving the original facial expression. This disentangled control over pose and expression spaces validates the effectiveness of our architecture.

\end{document}